\documentclass[letterpaper, 10 pt, conference]{ieeeconf}  

\IEEEoverridecommandlockouts                              

\usepackage{graphicx} 
\usepackage{subcaption}
\usepackage{siunitx}
\usepackage{amsmath} 
\usepackage{amssymb}  
\usepackage{algorithm}
\usepackage{algpseudocode}
\usepackage{booktabs}
\usepackage{tabularx}
\usepackage{xcolor}
\usepackage[normalem]{ulem}

\newcommand{\blue}[1]{#1}
\newcommand{\edit}[1]{#1}

\newcommand{\figref}[1]{Figure \ref{#1}}

\newcommand{\tableref}[1]{Table \ref{#1}}
\newcommand{\equationref}[1]{Equation~\ref{#1}}

\usepackage{cite}
\usepackage{multirow}

\title{\LARGE \bf
CueLearner: Bootstrapping and local policy adaptation from relative feedback}

\author{}

\author{Giulio Schiavi$^{1}$, Andrei Cramariuc$^{2}$, Lionel Ott$^{1}$, Roland Siegwart$^{1}$
\thanks{$^{1}$Autonomous Systems Lab, ETH Zurich, Switzerland}%
\thanks{$^{2}$Robotics Systems Lab, ETH Zurich, Switzerland.}%
\thanks{Correspondence: gschiavi@ethz.ch}%
\thanks{This work was partially supported by the HILTI Group.}%
\thanks{We wish to thank Pascal Roth (RSL, ETH Zurich) for his help in the setup of the Orbit simulation, and Helen Oleynikova (ASL, ETH Zurich) for her help in reviewing the manuscript.}%
}

\begin{document}

\maketitle
\thispagestyle{empty}
\pagestyle{empty}

\begin{abstract}
Human guidance has emerged as a powerful tool for enhancing reinforcement learning (RL). However, conventional forms of guidance such as demonstrations or binary scalar feedback can be challenging to collect or have low information content, motivating the exploration of other forms of human input. Among these, relative feedback (i.e., feedback on how to improve an action, such as “more to the left”) offers a good balance between usability and information richness. Previous research has shown that relative feedback can be used to enhance policy search methods. However, these efforts have been limited to specific policy classes and use feedback inefficiently. In this work, we introduce a novel method to learn from relative feedback and combine it with off-policy reinforcement learning. Through evaluations on two sparse-reward tasks, we demonstrate our method can be used to improve the sample efficiency of reinforcement learning by guiding its exploration process. Additionally, we show it can adapt a policy to changes in the environment or the user's preferences. Finally, we demonstrate real-world applicability by employing our approach to learn a navigation policy in a sparse reward setting.
\end{abstract}

\section{INTRODUCTION}
\label{sec:intro}
While reinforcement learning (RL) has achieved impressive results in multiple domains, it still suffers from several fundamental challenges, particularly when dealing with sparse or non-informative rewards. To address these difficulties, many works have introduced human guidance in the form of demonstrations~\cite{Levine_rl_manipulation_w_demonstrations,  deep_q_demos_2017}, action advice~\cite{Ilhan2021ActionAW}, binary scalar feedback~\cite{arakawa2018dqntamer, knox2010combining}, or preference feedback~\cite{2021pebble}. Compared with manually designing reward-shaping functions, these human-in-the-loop approaches have the potential to streamline the learning process and reduce engineering efforts. Each human input modality, however, presents unique trade-offs. For example, while demonstrations and action advice provide rich guidance to the agent, they are often challenging to produce as they require an expert and teleoperation infrastructure. Conversely, binary scalar scores and preferences are much easier to implement, but they convey less information and thus necessitate larger amounts of data.

An input modality that has, until now, received comparatively little attention is relative feedback. Relative feedback consists of guidance that suggests an improvement direction for the last action rather than dictating exact actions (e.g., “more to the left”). Compared with demonstrations and scalar/preference feedback, relative feedback offers a compromise: it does not require an expert trainer to provide the correct action, but still indicates a clear improvement direction for the agent, essentially providing a gradient in the action space. This could make it ideal for cases in which we want to guide RL exploration, or finetune a policy a posteriori, but do not have access to demonstrations or we expect them to be imprecise. \blue{Previous works have explored using relative feedback in combination with behavior cloning~\cite{PrezDattari2018InteractiveLW, Celemin2019ReinforcementLO}, or to bias a policy during deployment~\cite{cui2023totheright, shi2024yell}. However, current methods which apply relative feedback to reiforcement learning~\cite{Celemin2019ReinforcementLO} are not comparible with deep RL, and use human labels inefficiently.}

\blue{In this work, we propose a novel method to integrate relative feedback with off-policy deep reinforcement learning. Our method works by learning a feedback model from human labels, which is then used to refine the actions proposed by the agent. We use this to steer the agent's exploration during training or to refine its actions during deployment, allowing it to better adapt to changes in the environment. In constrast with other approaches that learn a feedback model~\cite{Celemin2019COACH, Celemin2019ReinforcementLO}, ours is compatible with deep RL. Moreover, our feedback model can fully replace the human trainer, allowing us to generalize a small number of annotations to novel states.} Our contributions are:
\begin{enumerate}
    \item \blue{We propose a novel method to integrate relative feedback with off-policy reinforcement learning.}
    \item \blue{We show that our approach effectively guides exploration in sparse reward tasks with a limited number of human annotations, and compare its label efficiency against other methods of guiding exploration which are found in the literature.}
    \item  We show that our approach can also be used to adapt existing policies, achieving similar label efficiency as learning a residual from optimal corrections while requiring less precise annotations.
    \item Finally, we demonstrate real-life applicability by using our method to learn a navigation policy in a sparse-reward setting, and deploying it in the real world.
\end{enumerate}

\section{RELATED WORK}
In this section, we discuss prior work on reinforcement learning with human feedback, human-in-the-loop reinforcement learning, and methods that employ relative feedback.

\subsection{Reinforcement Learning with Human Input}
Deep reinforcement learning, if performed without policy priors, requires many iterations to explore the solution space sufficiently. In addition, it is often challenging to explicitly define a sufficiently informative reward function, either because the task is inherently ambiguous or because, in real-world training scenarios, the agent does not have access to privileged information that might otherwise simplify the reward design. To sidestep this problem, a large number of works~\cite{ng_irl_2000, 2009tamer, 2017coach, 2021pebble, christiano_preferences, Hejna2022FewShotPL} assume no access to a traditional explicit environment reward and instead, attempt to learn a reward function from human input. In TAMER (Training an Agent Manually via Evaluative Reinforcement)~\cite{2009tamer}, the agent receives a binary scalar reward from a human trainer and then attempts to estimate the trainer’s reward function by assuming the provided feedback is a point estimate~\cite{2017coach}. Feedback in the form of preferences has also gained popularity, as human trainers find making relative comparisons easier than giving absolute judgments~\cite{2021pebble}. These methods estimate and refine the reward function by iteratively querying preferences between pairs of trajectory segments~\cite{christiano_preferences, 2021pebble, Hejna2022FewShotPL}. Finally, Inverse Reinforcement Learning methods~\cite{ng_irl_2000} attempt to estimate the environment reward from a set of task demonstrations. The main drawback of all the aforementioned approaches is that they can require a large number of human annotations to estimate a robust and unambiguous reward function.

\subsection{Human-in-the-Loop Reinforcement Learning}
A more general approach is to allow the agent to use both feedback from a trainer and from the environment. In these cases, human input can still be used to guide the learning process or add certain biases, while access to the environment reward reduces the need for human data. For example, a large family of works uses demonstrations or binary scalar feedback to provide an additional reward signal ~\cite{Levine_rl_manipulation_w_demonstrations, Goecks2019IntegratingBC, arakawa2018dqntamer}. \blue{Other methods leverage the ability of off-policy RL algorithms to learn from data beyond what the agent collects itself. This makes it possible to integrate a human teacher who assists the agent, by steering it toward valuable experiences or critical transitions. For example, this teacher can provide initial demonstrations~\cite{deep_q_demos_2017} or offer action advice for specific states~\cite{cruz_advising_2017}. However, directly using human input in this way can be inefficient. Instead, a more scalable alternative is to train an auxiliary policy that imitates the human teacher, which can then generalize to novel states without human supervision. This approach has been successfully applied to demonstrations~\cite{hansen2022modem}, action advice~\cite{Ilhan2021ActionAW} and scalar feedback~\cite{arakawa2018dqntamer, knox2010combining}. Our method is similar to those in this latter family. Compared with demonstration-based or action-advice methods, relative feedback reduces trainer effort by removing the need to provide optimal actions. Instead, it uses simple multiple-choice inputs like scalar feedback, but offers richer guidance by indicating how to improve actions. This makes it a promising middle ground—more informative than scalar feedback while remaining easy to use.}

\subsection{Learning with Relative Feedback} 
Several works have explored relative feedback—also termed “relative correction” or “directional feedback”\cite{Celemin2019ReinforcementLO, Celemin2019COACH, liu2023interactive, shi2024yell}. \blue{For instance, Shi et al.\cite{shi2024yell} bias behavior cloning (BC) policies with natural language instructions (e.g., “tilt your hand more to the left”), though this approach is limited to test-time adaptation and requires specific training of the BC policy. The work most similar to ours is that by Celemin et al.\cite{Celemin2019COACH}. These authors demonstrate using relative feedback to update the parameters of a policy parametrized as the linear combination of basis functions. The human trainer supplies the direction of the error between the action executed by the agent and the desired action, while the magnitude of the error is inferred by a linear model trained on the feedback history for each state. The same authors further extend their work by combining feedback-induced policy updates with task rewards in an actor-only policy search approach~\cite{Celemin2019ReinforcementLO}. 
This approach provides a solid foundation, but is not without limitations. The method by Celemin et al. is not compatible with deep reinforcement learning~\cite{Celemin2019ReinforcementLO}, which restricts it application to relatively simple tasks. Moreover, the human trainer is always required to supply the error direction, limiting the number of distinct states for which we can use human guidance. In contrast, our method is fully compatible with modern deep RL methods, and can tackle complex tasks with high-dimensional state-spaces. Moreover, our proposed feedback model can infer both the direction and magnitude of the action error. As a result, we can generalize feedback to novel states without further querying the human trainer.}

\section{PROBLEM SETTING}
We consider an agent that operates in an environment modeled as a Markov Decision Process (MDP) defined by the tuple $(S, A, P, R)$, where $S$ and $A$ refer to the continuous state and action spaces respectively, $P: S \times A \times S \rightarrow [0,1]$ is the state transition probability function, and $R: S \times A \rightarrow \mathbb{R}$ is the reward function.
At each timestep $t$ the agent executes an action $a \in A$ given a state observation $s \in S$ and receives a scalar reward $R(s,a)$. The \edit{action-value} function $Q: S \times A \rightarrow \mathbb{R}$ is defined as:
\begin{equation}
Q(s, a) = \mathbb{E} \left[ \sum_{t=0}^{\infty} \gamma^t R(s_t, a_t) \mid s_0 = s, a_0 = a \right]
\label{eq:cumulative_reward}
\end{equation}
where $\gamma \in [0,1]$ is the discount factor. After each action, the agent can query a trainer for feedback in the form of a normalized gradient $h$ in action space (i.e. the relative feedback/improvement direction):
\begin{equation}
h \in \nabla A = 
\begin{cases} 
\frac{\partial Q}{\partial a} \bigg/ \left\|\frac{\partial Q}{\partial a}\right\|_2 & \text{if $a$ is nonoptimal} \\
\mathbf{0} & \text{otherwise}
\end{cases}
\label{eq:relative_feedback}
\end{equation}

\section{METHOD}
\blue{Our method works by learning a feedback model to approximate the human annotator. We then use it to guide exploration in off-policy reinforcement learning or to refine a trained policy. In this section, we will first address how we train the feedback model. Then, we will explain how we integrate it in our two use-cases.}

\subsection{Learning a feedback model}
\label{sec:learning_rel_feedback}
Given an \blue{agent acting according to an} policy \(\pi_B(s)\) \blue{(which we will call the “base policy”)}, we collect \(N\) tuples of relative feedback \((s, a, h)\) from a trainer. We learn a model of the feedback $\phi_{\theta}$, where $\phi_{\theta}$ is a deep neural network parametrized by \(\theta\), by optimizing the following L2 loss: 
\begin{equation}
L(\theta) = \frac{1}{N} \sum_{i=1}^{N} \left\| \phi_{\theta}(s_i, a_i) - \epsilon \cdot h_i \right\|^2
\label{eq:loss_function}
\end{equation}
where $\epsilon$ is a hyperparameter which controls the strength of the feedback. 
Once we have learned the feedback model \(\phi_{\theta}\), we can apply it iteratively \blue{to refine the actions of the agent} without needing to query the human trainer again. Formally, let \( a_{0} = \pi_{B}(s) \) be the base policy’s action for a given state \( s \). We define the sequence
\begin{equation}
    a_{k+1} = \edit{a_{k}} + \phi_{\theta}\bigl(s,\,a_{k}\bigr),
    \quad k=0,1,2,\dots
    \label{eq:iterative_walk}
\end{equation} and continue refining the proposed action until one of the following conditions is met:
\begin{enumerate}
    \item The L2 distance between successive refinements is below a threshold $\tau$:
    \begin{equation}
      \|\,a_{k+1} - a_{k}\|_2 \;<\; \tau
      \label{eq:refinement_stopping_condition}
    \end{equation}
    \item A maximum number of iterations $K_{\max}$ is reached.
\end{enumerate}
\edit{where $\tau$ and $K_{\max}$ are task-dependent tuning parameters.} We define the refined action \edit{\(a_{\text{refined}} = a_{K}\)}
where $K \le K_{\max}$ is the iteration at which the sequence at \equationref{eq:iterative_walk} stops. In practice, this iterative refinement procedure suffers from compounding errors. \blue{The distribution of the state-action pairs for which we have feedback data depends on the base policy $\pi_B$. Each refinement step shifts actions away from this distribution, making the predictions on the model less reliable. We address this into two ways. The first is by} iteratively retraining the feedback model via Dataset Aggregation~\cite{2010dagger}, collecting new feedback for the refined actions at each iteration. In addition, if we have prior knowledge that the optimal action lies close to the base policy’s proposal \(a_{0}\), we can \blue{bound} the L2 distance between the refined action and \(a_{0}\):
\begin{equation}
    \bigl\lVert a_{\text{refined}} - a_{0} \bigr\rVert_{2} \;\le\; \tau_{\max},
    \label{eq:refined_action_bounded}
\end{equation}
and enforce this by \edit{clipping each action in the refinement sequence defined in~\equationref{eq:iterative_walk} to be within the ball of radius \(\tau_{\max}\) around \(a_{0}\).}

\subsection{Guiding exploration}
All reinforcement learning (RL) methods struggle with sparse rewards, where informative feedback signals are infrequent and hard to discover. However, off-policy reinforcement learning methods offer a distinct advantage: they do not require training exclusively on on-policy data. We can therefore leverage relative feedback to guide the RL agent and gather more informative transitions \edit{which are then included in} the replay buffer. We first train a feedback model \(\phi_{\theta}\) based on labels from a trainer (see previous section). To ensure good coverage of the action space we use a uniformly random base policy during this process:
\begin{equation}
    \pi_{B}(a \mid s) 
    =
    \text{Uniform}(A),
    \quad
    a \in A.
    \label{eq:uniform_base}
\end{equation}
We then use the learned feedback model to refine the agent’s actions during RL training\blue{, guiding it towards more informative transitions which are added to the replay buffer.} We empirically find that providing guidance every second episode balances the agent’s own exploration with feedback-guided exploration. Formally, on such guided episodes, we obtain an initial action from the RL policy,
\begin{equation}
    a_{0} = \pi_{RL}(s),
    \label{eq:initial_action_RL}
\end{equation}
and apply \blue{the feedback model} to obtain the refined action $a_{\text{refined}}$. The refined action is then executed in the environment. We cut off guidance after a predetermined number of episodes. \blue{Compared with previous approaches using relative feedback to guide exploration in reinforcement learning~\cite{Celemin2019ReinforcementLO}, in this stage the feedback model fully substitues the human trainer. This makes it possible to generalize a handful of labels to new points in the state-action space, without burdening the human trainer.} 

\subsection{Post-hoc adaptation}
\blue{In the previous section, we have explained how we use a feedback model, trained on human labels, to guide the exploration of a reinforcement learning agent. However,} once a policy is learned and deployed, we might find that some corrections are necessary, either because the agent or environment changed, or because the user's preferences changed. Rather than re-learning the policy from scratch or finetuning the network weights, we can \blue{again use a similar strategy} to refine the policy's actions towards a desired behavior. \blue{Specifically, let the base policy be:}
\begin{equation}
    \pi_{B}(a \mid s) 
    =
    \pi_{\text{to adapt}}(s)
    \label{eq:uniform_base_adaptation}
\end{equation}
We collect relative feedback labels from a trainer and learn a feedback model, as described in Section~\ref{sec:learning_rel_feedback}. As previously mentioned, we train iteratively using feedback collected on the base policy, as well as feedback on the actions refined in subsequent iterations, which makes the feedback model robust to compounding errors. \blue{In the end, the feedback model learns to robustly refine the actions of the base policy, adapting them to our new preferences (we again use the iterative procedure in Section \ref{sec:learning_rel_feedback} to infer the refined action).} \blue{We note that, while previous methods~\cite{Celemin2019ReinforcementLO, Celemin2019COACH} use relative feedback to update the parameters of the policy directly, we instead learn a separate action refinement module. This makes our method agnostic to the parametrization of the base policy, increasing generality.}

\section{EXPERIMENTAL SETUP}
\begin{figure}[bt]
    \centering
    \vspace{5mm}
    \includegraphics[width=0.45\textwidth]{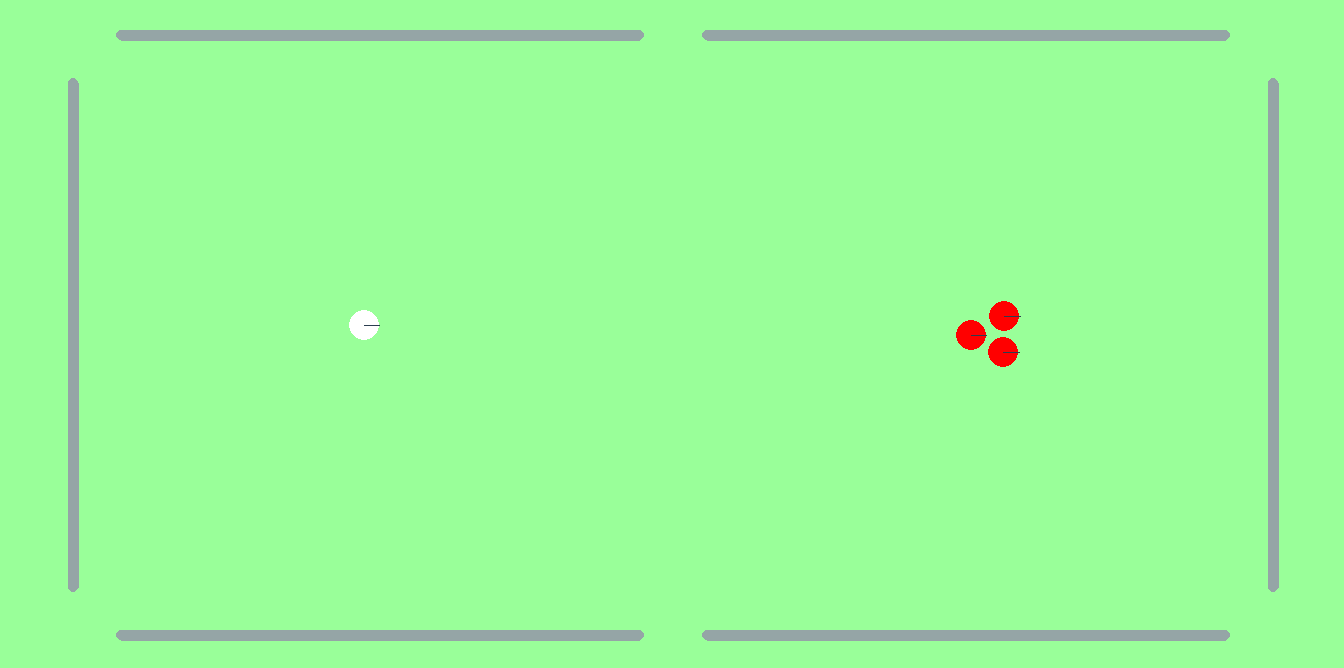}
    \caption{Setup and initial configuration of our billiards task. To prevent the agent from overfitting, we add a small Gaussian disturbance to the initial positions of the target balls.}
    \label{fig:billiards_table}
\end{figure}
\begin{figure}[bt]
    \centering
    \includegraphics[width=0.45\textwidth]{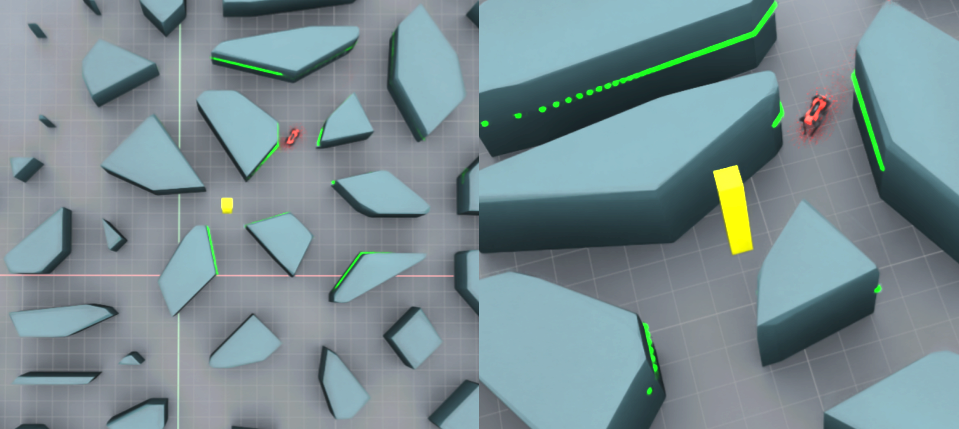}
    \caption{Top-view and side-view of our navigation environment in the Orbit simulator~\cite{mittal2023orbit}. The robot (red) navigates towards the goal position (yellow). The green dots visualize the lidar observation.}
    \label{fig:navigation_environment}
\end{figure}

We test our method on two sparse-reward tasks: a simplified billiards game and map-less navigation of a quadruped. Both tasks challenge conventional RL: in billiards, most actions do not lead to any reward, while in the quadruped navigation task, the reward is only provided upon reaching the goal, making it sparse and delayed. We deliberately avoid adding additional reward terms, to evaluate whether our method is effective even in extremely sparse reward scenarios. Furthermore, the two tasks vary significantly in their state spaces and dynamics, allowing us to assess our method's capacity to generalize across diverse problem domains. We conduct the majority of tests in simulation, but then validate our method through a real-world deployment.

\subsection{Billiards task}
We simulate a billiards table with three target balls and one cue ball, as illustrated in \figref{fig:billiards_table}. The goal is to pocket all target balls in as few moves as possible. At each step, the agent observes the positions of all balls and chooses the direction to strike the cue ball in \edit{(represented as a unit-vector in $\mathbb{R}^2$)}. It receives a reward of $1$ if a target ball was pocketed, $0$ otherwise. We measure policy performance in this task by the average step reward, corresponding to the probability of pocketing a ball at each step.

\subsection{Navigation task}
We simulate a quadruped robot (ANYmal~\cite{hutter_anymal}) in a procedurally-generated 3D environment. The task is to navigate to a goal without collisions. At each step, the robot receives a 2D lidar scan observation ($270^\circ$ f.o.v. and $1^\circ$ angular resolution), its position relative to the goal, and a vector containing its three previous actions. The agent then chooses a movement direction. \edit{The robot has access to a low-level locomotion controller, which takes as input the desired movement direction and outputs joint commands.} The goal is placed between 5 and 15 meters away. The agent receives a reward of $10$ for reaching the goal, $-10$ for a collision, and $0$ otherwise. A visualization of the environment is shown in \figref{fig:navigation_environment}. Performance is measured by the success rate and collision rate, which are respectively the percentages of episodes where the agent reaches the goal without timing out and those terminated due to a collision.

\subsection{Feedback generation}
\label{sec:feedback_generation}
Although our method is designed for human feedback, relying on human annotators for the required number of experiments and random seed runs would be impractical. Therefore, we implement oracles for both tasks to efficiently and repeatably query feedback labels (\edit{both demonstrations and relative feedback}). For the billiards task, we train an agent using our baseline (double deep Q-learning~\cite{van2016deep}). For the navigation task, we use an oracle based on probabilistic roadmaps~\cite{kavraki_probabilistic_roadmaps}. We nevertheless confirm the feasibility of our approach with actual human feedback in our final real-world navigation experiment.

\section{EXPERIMENTS}
In this section, we begin by evaluating how our method can bootstrap the exploration process of an RL agent, \blue{and compare it against alternative forms of human feedback.}
We then evaluate our method in the adaptation use case and compare its label efficiency with residual learning from optimal corrections. Finally, we demonstrate the real-world feasibility of our method by training a navigation policy with actual human feedback and deploying it on hardware. Throughout all experiments (with the exception of real-world deployment), we report the average validation performance across five independent policies, each initialized with a unique random seed.

\subsection{Guiding exploration}
\blue{Previous work has shown that off-policy RL can be guided by auxiliary teacher policies learned from demonstrations~\cite{hansen2022modem, Ilhan2021ActionAW} or binary scalar scores~\cite{knox2010combining}. Demonstrations are more label-efficient but can require expert teachers and teleoperation setups. Scalar feedback is easier to implement but less informative, needing more human annotations. Relative feedback offers a middle ground, requiring no demonstrations yet providing richer guidance than scalar feedback. We compare our method’s label efficiency with both approaches. We are particularly interested in the comparison with scalar feedback, since it requires a similar level of labeling effort. In contrast, demonstrations serve as an upper bound on performance, as each label conveys significantly more information. We also benchmark the behavior of the RL agent without guidance, as well as an agent where exploration is guided by a simple heuristic (e.g., hitting the closest ball or walking toward the goal). We do not compare our method with Celemin et al.~\cite{Celemin2019ReinforcementLO}, as their approach is not compatible with deep RL.} In all experiments, the RL agent is implemented as a double deep Q-learning (DDQN)~\cite{van2016deep} policy, which we choose for its ease of use. For the navigation task, we additionally use an n-step temporal difference update~\cite{sutton_rl} to allow the sparse reward signal to better propagate through the value function. \edit{We simulate human feedback using the oracles described in Section~\ref{sec:feedback_generation}}. We report hyperparameters in the Appendix~\ref{app:training_details}.

\begin{figure}[bt]
    \centering
    \includegraphics[width=0.5\textwidth]{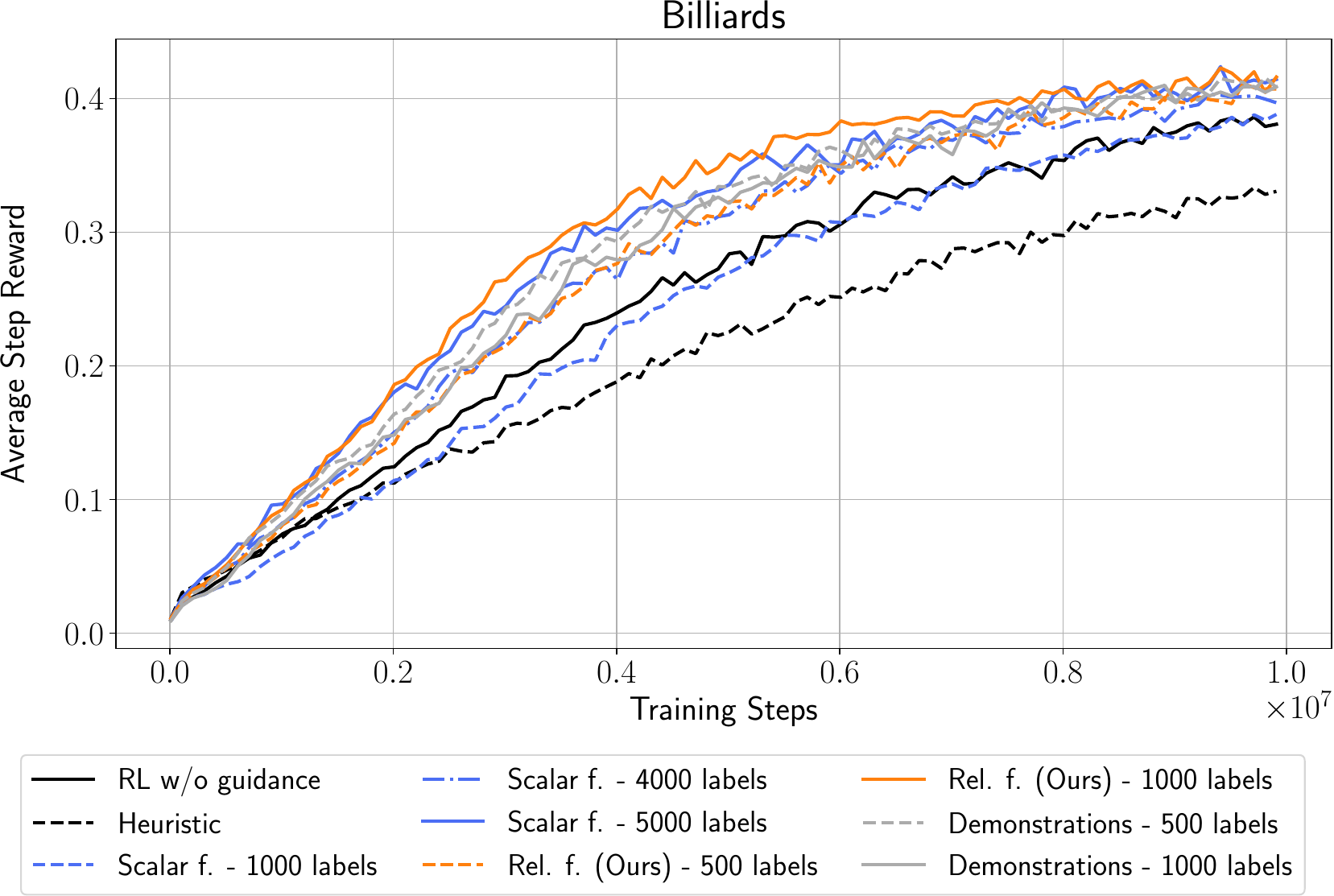}
    \caption{Average validation performance on the billiards task. Our approach outperforms RL with no guidance. It also slighly outperforms guidance with demonstrations with the same amount of labels, while scalar feedback requires five times more labels.}
    \label{fig:bootstrapping_billiards}
\end{figure}

\begin{figure}[bt]
    \centering
    \includegraphics[width=0.5\textwidth]{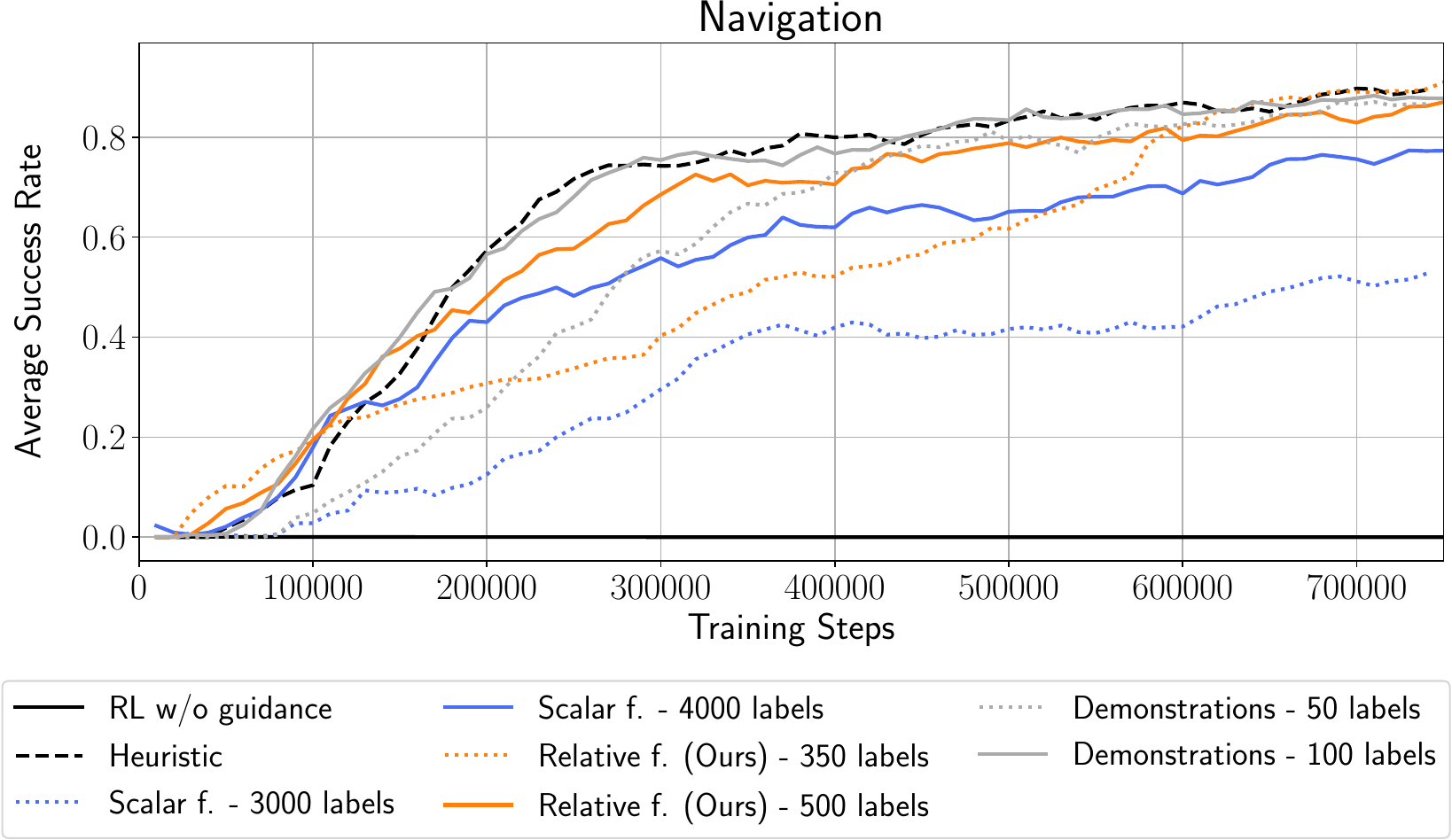}
    \caption{Average validation performance in the quadruped navigation task. In this task RL without guidance is ineffective due to the sparse reward. Guidance from demonstrations requires the smallest amount of labels, likely because only a few are needed to learn to go towards the goal. Our method using relative feedback is effective from 350 labels, while scalar feedback requires ten times more labels.}
    \label{fig:bootstrapping_navigation}
\end{figure}

Our results are summarized in~\figref{fig:bootstrapping_billiards} and~\figref{fig:bootstrapping_navigation}. \blue{We observe the billiards task is more difficult than navigation, requiring ten times more steps to converge.} In this task, we find that our method outperforms the pure RL agent using a moderate amount of relative feedback labels (500 to 1000), reducing the number of training steps needed by around a third. \blue{Compared with scalar feedback, our method requires five times less labels to achieve the same speedup. We also outperform using a heuristic, suggesting our feedback model can capture more complex or informative behavior. Interestingly, our method performs similarly to or slighly better than guidance from demonstrations.} In the navigation task, the sparse rewards make traditional RL ineffective. In contrast, our method successfully guides exploration with 350–500 relative feedback labels without additional reward shaping. \blue{Again, we significantly outperform scalar feedback, requiring almost ten times less labels to reach the same performance. In this task, demonstrations require far fewer labels (only 50 to 100). This is likely because a simple and imprecise strategy of heading towards the goal (which is a unimodal strategy easily learned from demonstrations) works very well. This is supported by the performance of the heuristic closely mirroring that of the demonstrations.}

\blue{In summary, we find that guiding exploration using relative feedback is significantly more label-efficient compared with scalar feedback, despite similar per-label complexity. The label requirement of our method is practical for actual usage (in a later experiment, collecting 500 relative feedback labels took us 40 minutes). Interestingly, while relative feedback requires more labels than demonstrations on the navigation task, it performs comparably in the more challenging billiards task, despite placing a lower burden on the trainer.} \edit{As a side note, from a computational standpoint our  method adds only minimal overhead compared to the pure RL baseline.}

\subsection{Post-hoc adaptation - Billiards}
\begin{figure}[bt]
    \centering
    \includegraphics[width=0.5\textwidth]{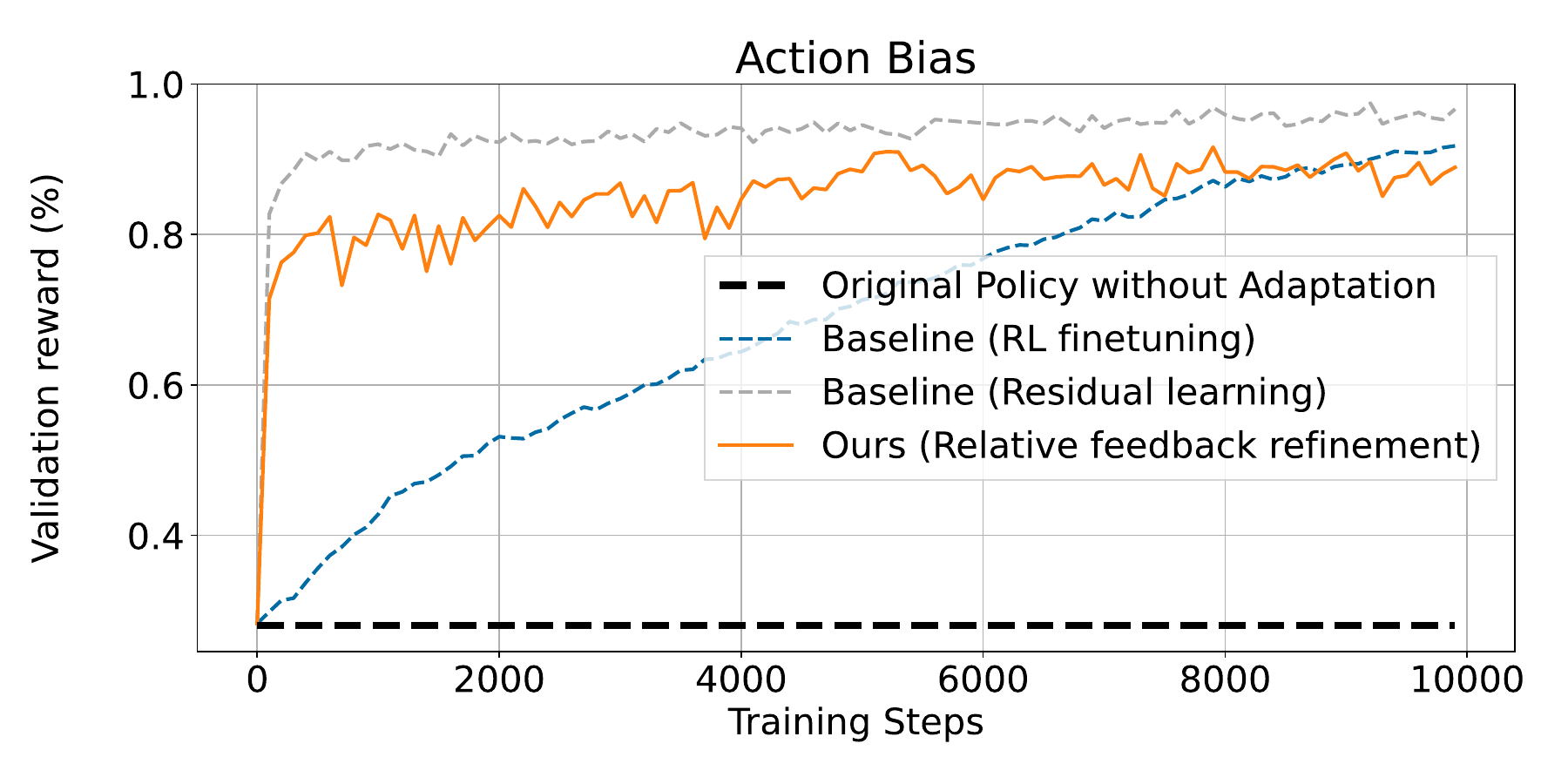}
    \includegraphics[width=0.5\textwidth]{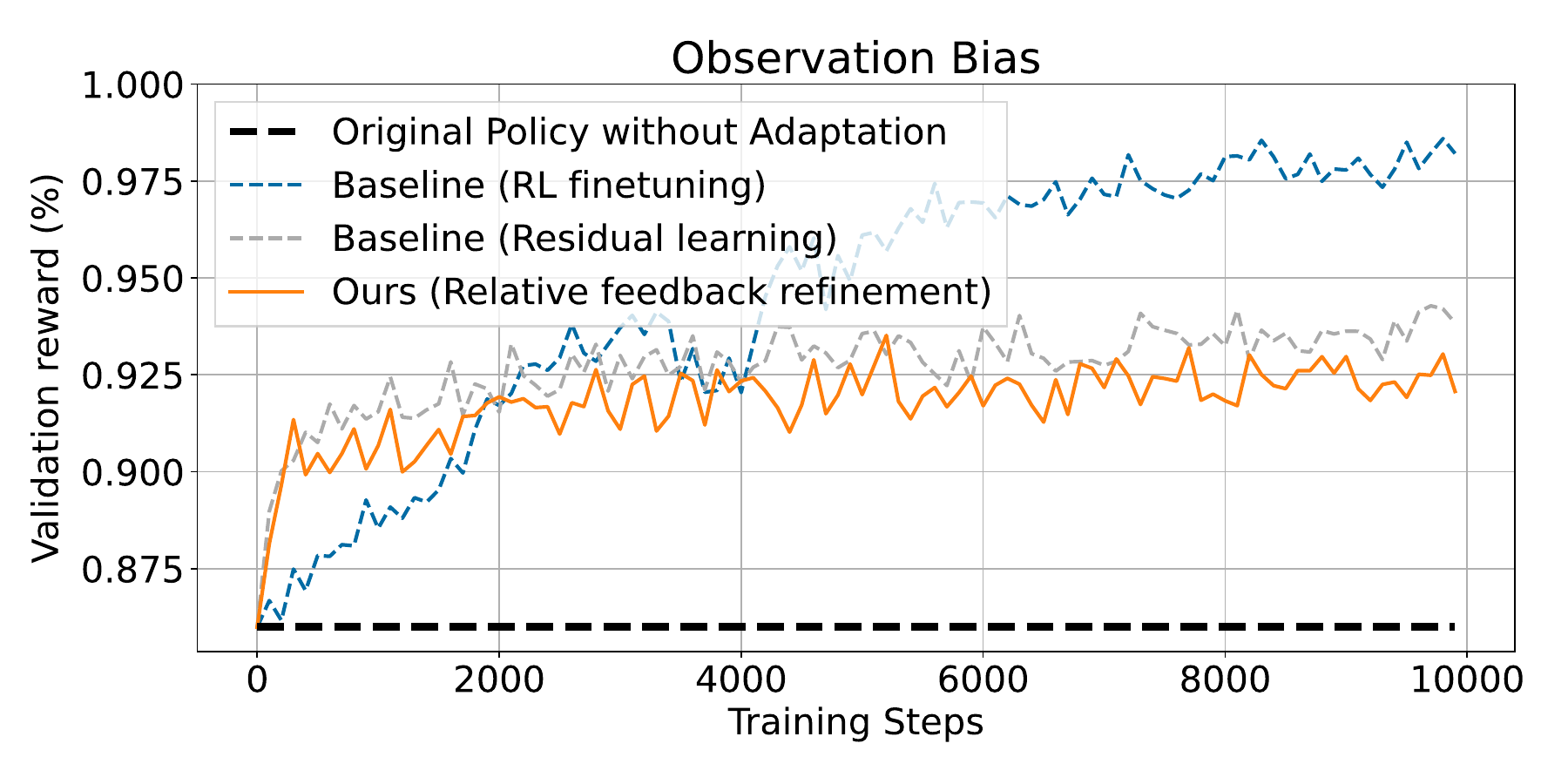}
    \includegraphics[width=0.5\textwidth]{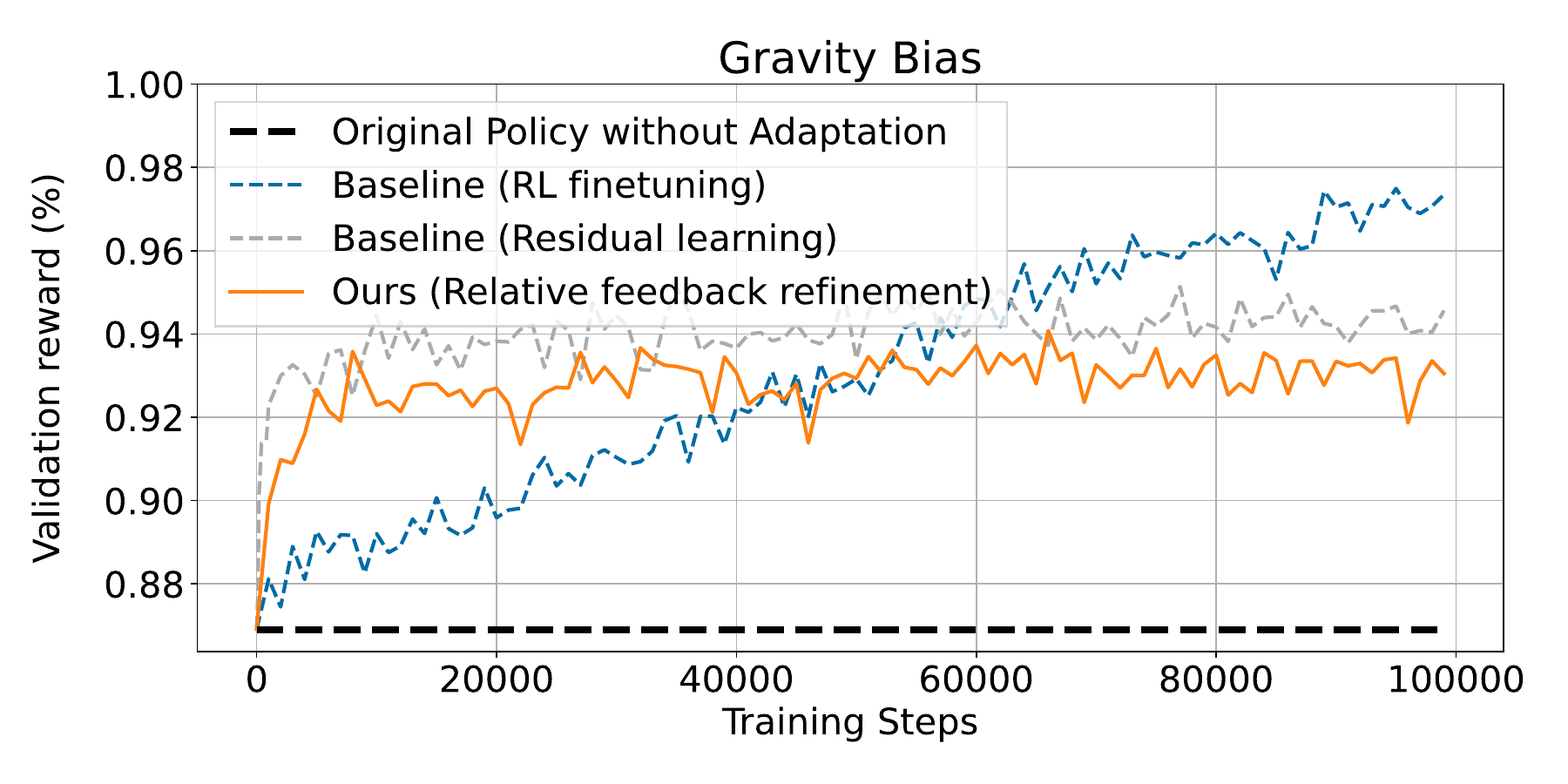}
    \captionsetup{aboveskip=-5pt}
    \caption{Comparison of four approaches -- no adaptation, RL finetuning, residual learning from optimal corrections, and our method -- on three billiard environment changes. Performance is expressed as a percentage of the original reward and averaged over five seeds. While our method does not fully recover the original performance, it adapts more rapidly than RL finetuning. It also achieves results on par with residual learning, indicating it is label-efficient and can reach the best possible result within the adaptation bounds.}
    \label{fig:adaptation_billiards}
\end{figure}

In the next set of experiments, we adapt the previously learned billiards and navigation policies to changes in the environment or in user preferences. We examine six different adaptation scenarios, three for each task. In the first three scenarios, we alter the billiards task in three ways: we simulate embodiment changes by adding a constant bias to each action, sensor calibration errors by adding a bias to the ball observations, and environment shifts by slightly changing the direction of gravity in the simulation. We compare our method to two baselines:
\begin{enumerate}
    \item Finetuning the RL policy on the new environment, which provides a lower bound on adaptation speed without human feedback.
    \item Learning a residual policy (\edit{i.e. learning to directly predict the needed correction}), which \edit{uses optimal corrective demonstrations and} serves as an upper bound on achievable performance. 
\end{enumerate}
For both our method and the residual learning baseline, we find that trying to infer a global residual results in instability. To address this, we impose bounds on the residual of $\pm\SI{1}{\degree}$ for the embodiment adaptation and of $\pm\SI{0.3}{\degree}$ for the other two scenarios (see \equationref{eq:refined_action_bounded}). Selecting these bounds requires some estimation, but we find they are intuitive to choose (as they correspond to meaningful physical quantities) and are robust to small variations. Our results are summarized in \figref{fig:adaptation_billiards}. Even though RL finetuning eventually recovers all lost performance, our method adapts around one order of magnitude faster. Notably, while we cannot fully restore the original performance due to the imposed bounds, our method achieves similar label efficiency and nearly the same final performance as the residual-based learning baseline, demonstrating our approach can efficiently reach the best achievable adaptation within the given constraints while requiring far less precise annotations.

\subsection{Post-hoc adaptation - Navigation}
\begin{figure}[bt]
    \centering
    \includegraphics[width=0.45\textwidth]{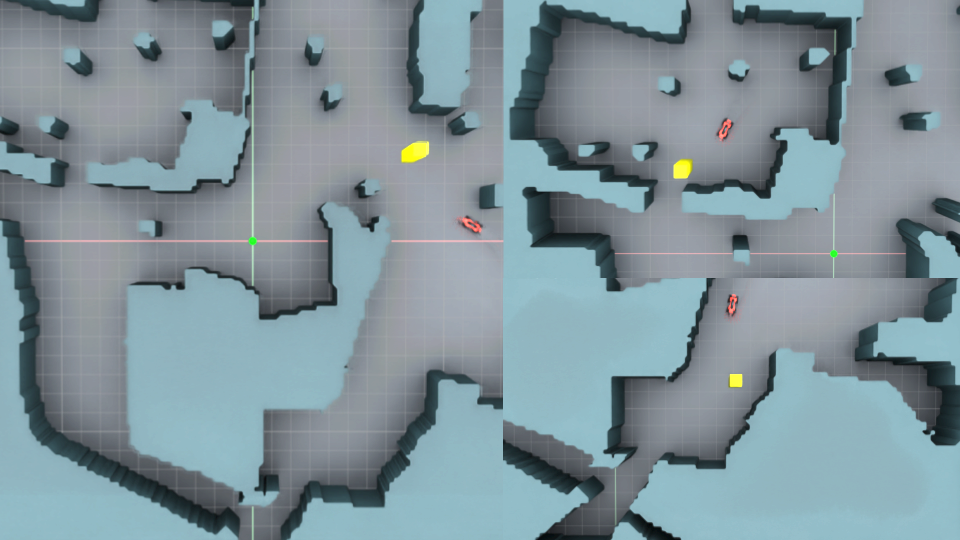}
    \caption{The environment used for our adaptation experiment. Compared with the procedurally generated training environments (\figref{fig:navigation_environment}) it is more diverse, presents thinner obstacles, rooms with doorways, and narrow corridors.}
    \label{fig:adaptation_navigation_new_enviornment}
\end{figure}

In the next two experiments we evaluate adaptation to environment changes in the navigation task. In the first, we simulate a small sensor miscalibration by rotating the lidar signal by a constant \SI{15}{\degree}. In the second, we test the limits of our method by attempting to adapt to an environment that differs substantially from the training conditions. The difference between the new and training environment can be seen by comparing Figures~\ref{fig:adaptation_navigation_new_enviornment} and~\ref{fig:navigation_environment}. 

In the first case, we find that 500 relative feedback labels are enough to restore a substantial portion of the policy's performance. For instance, introducing lidar rotation error reduces the policy’s success rate from $93\%$ to $72\%$, but our method improves it back to $82\%$. Note that, like in the billiards task, we use a bound of $\pm\SI{60}{\degree}$ on the adaptation to improve stability. Our results are summarized in \tableref{tab:navigation_adaptation_lidar}. Unfortunately, our approach struggles to improve performance in the second case, when we attempt to adapt to an environment that significantly deviates from the original training conditions. If we impose a bound of $\pm\SI{60}{\degree}$ on the residual, it merely preserves performance, and with a more generous limit (up to $\pm\SI{180}{\degree}$), it actually worsens the results. To understand why this is the case, we replace the learned feedback model with our probabilistic roadmap oracle, retaining the same bounds on action adjustments. The oracle also fails to improve performance when restricted to $\pm\SI{60}{\degree}$, while it does help when allowed to deviate actions by up to $\pm\SI{180}{\degree}$ (see \tableref{tab:navigation_adaptation_environment}). This implies that improving performance requires correcting the base policy’s actions by more than $\pm\SI{60}{\degree}$. These results suggest that our method is well-suited for making localized adjustments, but is less effective in scenarios that require extensive global modifications.

\begin{table}[bt]
\centering
\vspace{3mm}
\begin{tabular}{ll}
\toprule
\multirow{2}{*}{\textbf{Policy}} & \textbf{Success Rate} \\
 & \textbf{(Collision Rate)} \\
\hline
Original w/o lidar offset & 93\% (2\%) \\
Original w lidar offset & 72\% (24\%) \\
Human feedback refinement w lidar offset & 82\% (15\%)\\
\bottomrule
\end{tabular}
\caption{Comparison between the performance of the policy on the original environment it was trained on versus performance on the environment with the miscalibrated lidar, versus using relative feedback to compensate for the miscalibration. With a small number of labels (500), we can recover a significant amount of the lost performance.}
\label{tab:navigation_adaptation_lidar}
\end{table}

\begin{table}[bt]
\centering
\begin{tabular}{ll}
\toprule
\multirow{2}{*}{\textbf{Policy}} & \textbf{Success Rate} \\
 & \textbf{(Collision Rate)} \\
\midrule
Original w/o adaptation & 68\% (4\%) \\
Human feedback refinement w $\pm\SI{60}{\degree}$ bound & 65\% (8\%) \\
Human feedback refinement w $\pm\SI{180}{\degree}$ bound & 61\% (21\%) \\
PRM oracle refinement w $\pm\SI{60}{\degree}$ bound & 73\% (3\%) \\
PRM oracle refinement w $\pm\SI{180}{\degree}$ bound & 93\% (3\%) \\
\bottomrule
\end{tabular}
\caption{Performance comparison of adaptation strategies in the new environment. Our method (human feedback refinement) preserves performance under a $\pm\SI{60}{\degree}$ action refinement limit but worsens it under $\pm\SI{180}{\degree}$. Meanwhile, the probabilistic roadmap (PRM) oracle -- an upper bound on achievable adaptation -- only improves performance under a $\pm\SI{180}{\degree}$ bound. This indicates that this scenario requires global action adaptation, which our method is not suited to learn.}
\label{tab:navigation_adaptation_environment}
\end{table}

\begin{figure}[bt]
    \centering
    \vspace{3mm}
    \includegraphics[width=0.3\textwidth]{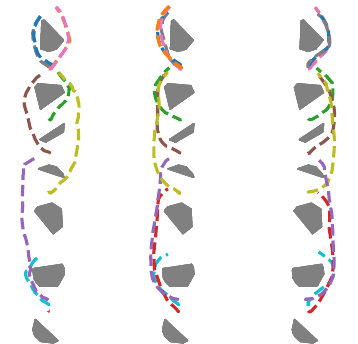}
    \caption{Qualitative trajectory comparisons illustrating the adaptation of policies to user preferences using relative feedback. Left: original policy. Center: adapted to avoid obtacles on the right. Left: adapted to prefer avoidance on the left.}
    \label{fig:adaptation_avoidance_behavior}
\end{figure}

In our last simulation experiment, we demonstrate that our method can use relative feedback to enforce user preferences. We deploy the learned navigation policy in a corridor-like environment with obstacles and we provide feedback to consistently favor obstacle avoidance on a preferred side. With around 500 feedback labels, the policy reliably avoids obstacles on the chosen side (right or left) without reducing overall navigation success. The success rate of the original policy is $80\%$ and avoids obstacles to the right or left with equal probability ($48\%$ each, with of $4\%$ “mixed” cases). After incorporating the feedback, the success rate of the policy remains high, averaging at $83\%$ (average of “left” and “right” policies). The updated policies demonstrate a strong alignment with the user preferences, avoiding obstacles to the preferred side (either right or left) in $96\%$ of cases. We show some qualitative trajectory comparisons in~\figref{fig:adaptation_avoidance_behavior}. This suggests that, in simple scenarios, our method can be used to enforce user preferences without leading to a drop in policy performance.

\subsection{Real-world experiment}
\begin{figure}[bt]
    \centering
    \vspace{3mm}
    \includegraphics[width=0.45\textwidth]{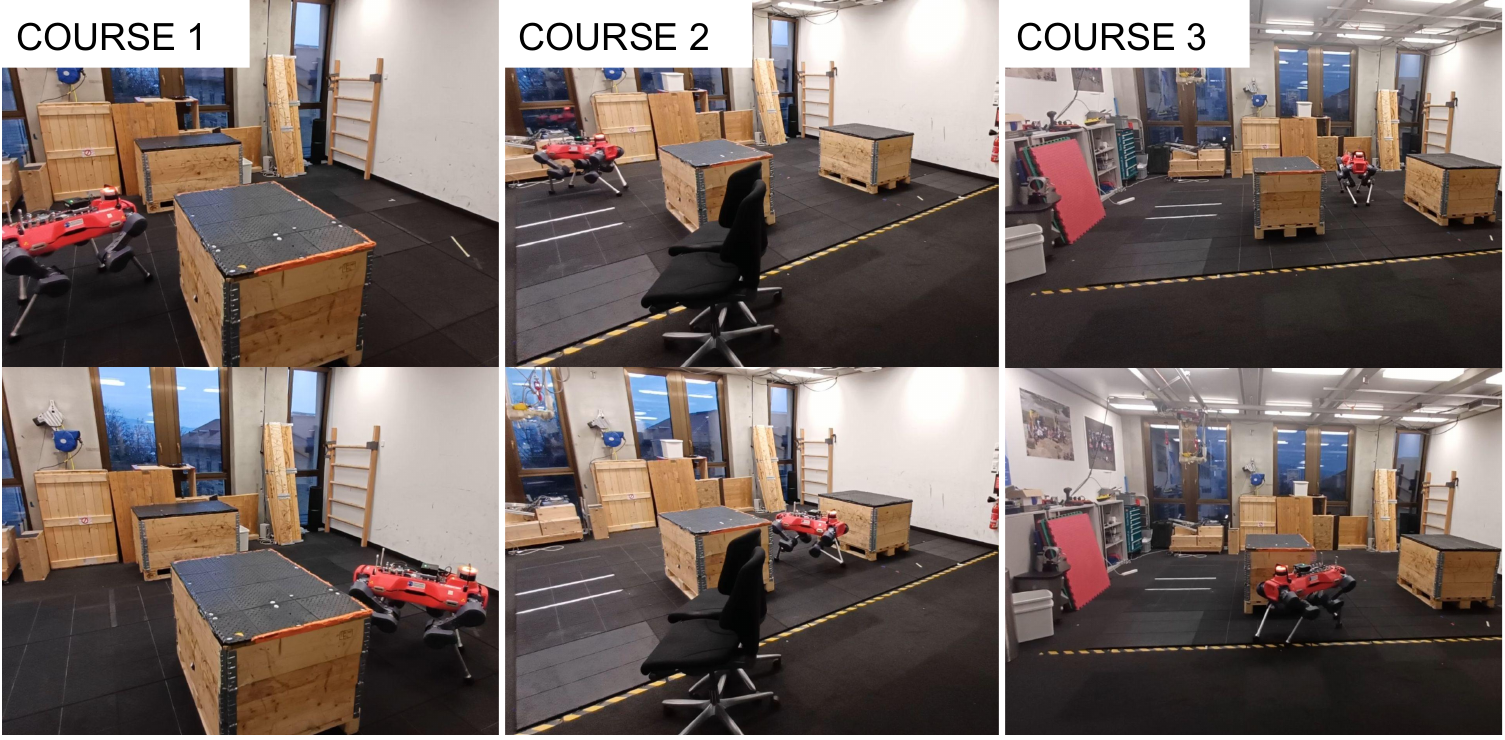}
    \caption{We use our policy to navigate an ANYmal though several obstacle courses, of which we show several examples in this figure. Videos of the tests are available in the supplementary.}
    \label{fig:real_world_navigation}
\end{figure}

In this final experiment, we show our method enables a real-world practitioner to bootstrap a navigation policy through feedback and deploy it on a real robot. Instead of relying on an oracle, we provide the feedback ourselves via a multiple choice interface. We select whether the current action:
\begin{enumerate}
    \item Is already optimal.
    \item Should be adjusted counterclockwise.
    \item Should be adjusted clockwise.
\end{enumerate}
We are able to collect 500 human labels in less than 40 minutes, and these are more than sufficient to bootstrap policy learning. We then deploy the resulting policy, navigating an ANYmal robot through a real-world obstacle course, as depicted in~\figref{fig:real_world_navigation}. Our qualitative observations show that, despite being trained entirely in simulation without any additional sensor calibration or sim-to-real adaptations, the policy transfers effectively to the real world. It handles sensor noise well, traverses obstacle courses with minimal collisions, and experiences only minor scrapes, which arise from discrepancies between the real robot’s physical geometry and the simplified collision bodies used in simulation. This provides strong evidence that our training method produces high-quality policies without compromising performance.

\section{CONCLUSIONS}
\blue{In this work, we tackled the problem of robotic learning using relative feedback. We introduced a novel method to learn a relative feedback model from human labels. We showed this feedback model can be integrated with off-policy RL, to improve sample-efficiency in sparse reward scenarios by guiding exploration. Through simulation experiments, we showed our method requires a practical amount of human annotations to be effective. We showed our method is significantly more label-efficient than using scalar feedback, and on some tasks perfoms comparably or slighly better than using guidance from demonstrations. We additionally showed our method can be further used during deployment to adapt the policy. In this case, our method is significantly more data-efficient than finetuning the RL policy without guidance, and has comparable label efficiency to residual learning from optimal demonstrations. Finally, we showed our method is straightforward to apply in real-world training scenarios, requiring minimal effort to produce a policy that is ready for deployment.} We believe its ease of implementation, combined with its moderate-to-low data requirements and integration with reinforcement learning, makes our method an interesting new tool for robot learning, particularly in cases where access to optimal demonstrations from an expert is limited.






\section*{Acknowledgments}
We acknowledge the use of artificial-intelligence-based tools as writing aids (editing and grammar enhancement) in the preparation of this manuscript. All intellectual content and conclusions remain the sole responsibility of the authors.

\bibliographystyle{IEEEtran}
\bibliography{references}

\appendix
\label{app:training_details}
\paragraph{Billiards Task} 
We train the DDQN policy using an MLP with layers [400, 400, 400, 300, 300, 300], a 1:1 gradient-to-environment step ratio, an SGD optimizer (learning rate 0.1), and a discount factor of 0.0. The feedback model is an MLP with layers [400, 400, 300, 300], trained from scratch \edit{every 10 new feedback labels} using Adam (learning rate 0.001). The feedback strength hyperparameter is set to 
\(\epsilon = 5^\circ\) for guided exploration, and \(\epsilon = 0.1^\circ\) for post-hoc adaptation. For demonstration and scalar feedback baselines, the auxiliary guidance policy uses the same hyperparameters as the relative feedback model. For the scalar feedback baseline, we assign a positive label to any action within \(5^\circ\) of the correct one, as this yields the best performance.

\paragraph{Navigation Task} We train the DDQN policy using an MLP with layers [400, 400, 400, 300, 300, 300], a 1:1 gradient-to-environment step ratio, an SGD optimizer (learning rate 0.001), and a discount factor of 0.995. The feedback model is shared with the billiards task. The feedback strength hyperparameter is $\epsilon=\SI{20}{\degree}$ for guided exploration experiments, and $\epsilon=\SI{2}{\degree}$ for the post-hoc adaptation experiments. For the scalar feedback baseline, we assign a positive label to all actions within $20^\circ$ of the correct action (which achieves the best performance).

\end{document}